\documentclass{article}

\usepackage[preprint]{neurips_2024}
\usepackage{natbib} 
\usepackage{graphicx}
\usepackage{amsmath}
\usepackage{multirow}
\usepackage{booktabs}
\usepackage{caption}
\usepackage{adjustbox}
\usepackage{setspace}
\usepackage{subfigure}
\usepackage{booktabs} 
\usepackage{makecell}

\usepackage[utf8]{inputenc} 
\usepackage[T1]{fontenc}    
\usepackage{hyperref}       
\usepackage{url}            
\usepackage{booktabs}       
\usepackage{scrextend}
\usepackage{amsfonts}       
\usepackage{nicefrac}       
\usepackage{microtype}      
\usepackage{xcolor}         
\usepackage{graphicx}
\usepackage{enumitem}
\usepackage[capitalise]{cleveref}
\usepackage{comment}
\usepackage{color-edits}
\addauthor{cm}{blue}
\addauthor{todo}{red}
\usepackage{multirow}
\usepackage{xcolor}         
\usepackage{float}

\title{Sparse Matrix in Large Language Model Fine-tuning}

\author{%
  Haoze He\thanks{Equal Contribution} \\
  School of Computer Science\\
  Carnegie Mellon University\\
  Pittsburgh, PA 15213 \\
  \texttt{haozeh@cs.cmu.edu} \\
  \And
  Juncheng Billy Li \footnotemark[1] \\
  School of Computer Science \\
  Carnegie Mellon University\\
  Pittsburgh, PA 15213  \\
  \texttt{junchenl@cs.cmu.edu} \\
  \AND
  Xuan Jiang \\
  College of Engineering \\
  University of California, Berkeley \\
  Berkeley, CA 94720 \\
  \texttt{xuanjiang@berkeley.edu} \\
  \And
  Heather Miller\thanks{Corresponding Author} \\
  School of Computer Science \\
  Carnegie Mellon University \\
  Pittsburgh, PA 15213  \\
  \texttt{heather.miller@cs.cmu.edu} \\
}

\begin{document}
\maketitle

\begin{abstract}
    LoRA and its variants have become popular parameter-efficient fine-tuning (PEFT) methods due to their ability to avoid excessive computational costs.
    However, an accuracy gap often exists between PEFT methods and full fine-tuning (FT), and this gap has yet to be systematically studied.
    In this work, we introduce a method for selecting sparse sub-matrices that aim to minimize the performance gap between PEFT vs. full fine-tuning (FT) while also reducing both fine-tuning computational cost and memory cost.
    Our Sparse Matrix Tuning (SMT) method begins by identifying the most significant sub-matrices in the gradient update, updating only these blocks during the fine-tuning process.
    In our experiments, we demonstrate that SMT consistently surpasses other PEFT baseline (e.g. LoRA and DoRA) in fine-tuning popular large language models such as LLaMA across a broad spectrum of tasks, while reducing the GPU memory footprint by 67\% compared to FT.
    We also examine how the performance of LoRA and DoRA tends to plateau and decline as the number of trainable parameters increases, in contrast, our SMT method does not suffer from such issue.
\end{abstract}

\section{Introduction}

While the inherent generalization capability of Large Language Models (LLMs) is impressive, enhancing performance on downstream tasks often still necessitates fine-tuning \cite{ding2022delta, chung2022scaling}.
However, as the size of these LLMs increases, there is a pressing challenge to optimize the fine-tuning process for better computational efficiency and memory utilization.
For example, fine-tuning a pre-trained LLaMA 7B model without CPU offloading\footnote{Although some libraries such as Deepspeed can move the optimizer memory cost to CPU, it will also slow down the fine-tuning with extra I/O communication time \cite{rajbhandari2020zero, aminabadi2022deepspeed}.}
requires at least 58 GB of GPU vRAM---13.6 GB for trainable parameters, 40 GB for Adam optimizer states and gradients, and 2 GB for activation. 
This requirement makes fine-tuning on consumer-level GPUs such as the NVIDIA RTX 4090 with 24 GB of memory impractical \cite{zhao2024galore}.

To address the prohibitive computational challenges of full parameter fine-tuning, many parameter-efficient fine-tuning (PEFT) methods have emerged over the past two years. LoRA and its variants \cite{hu2021lora, zhao2024galore, dettmers2024qlora, liu2024alora, liu2024dora} utilize the weight low-rank adaptation method and successfully reduce both the optimizer memory and computational cost.
However, even in state-of-the-art (SoTA) PEFT research, results show a notable performance gap between low-rank adaptation methods and full parameter tuning across many datasets \cite{liu2024dora}.
Additionally, in this work, we report a less recognized phenomenon: low-rank adaptation PEFT methods experience a performance plateau and subsequent decline as the parameter count (rank r) increases. Surprisingly, even with more trainable parameters, performance decreases.

On the other hand, previous studies have extensively analyzed the internal logic of LLMs. Some knowledge editing methods, such as Constrained fine-tuning \cite{zhu2020modifying}, ROME \cite{meng2022locating}, and MEMIT \cite{meng2022mass}, have shown that LLMs have memory sections located in distinct layers. These memories can be modified via fine-tuning \cite{zhu2020modifying}.
These works observed that domain-specific knowledge can be distributed separately and sparsely among layers.
Motivated by these observations, and to address the challenges of PEFT mentioned above, we proposed a \emph{\textbf{Sparse Matrix Tuning(SMT)}} approach.
By applying matrix sparsity,  we aim to identify and fine-tune the most relevant memory sections efficiently.
Contrary to \cite{geva2020transformer, geva2022transformer}'s claim that transformer's MLP layers primarily serve as key–value memories, we empirically demonstrate that attention mechanisms, particularly the value vector, store the largest number of memories and are the most influential during fine-tuning.

In our experiments, our Sparse Matrix Tuning (SMT) approach achieves better performance compared to LoRA and DoRA using same amount of trainable parameters.
Additionally, SMT closes the accuracy gap between full fine-tuning, overcomes the performance plateau of low-rank adaptation PEFT methods, and significantly outperforms LoRA and DoRA while utilizing less than 5\% of trainable parameters. 
Our experimental results show that SMT consistently outperforms SoTA PEFT (including LoRA and DoRA) methods by 2+ points when fine-tuning popular LLMs (e.g. LLaMA series base model\footnote{Base model is not yet instruction tuned}) on commonsense reasoning and arithmetic reasoning benchmarks.
Our experimental results show that SMT consistently outperforms DoRA, such as commonsense reasoning (+3.0/+2.8 on LLaMA-7B/13B,
+2.9 on LLaMA2-7B, and +2.0 on LLaMA3-8B) and arithmetic reasoning (+2.3 on LLaMA-7B).
In addition, SMT eliminates the accuracy gap between SMT and full fine-tuning, overcomes the plateau of low-rank adaption PEFT methods, and significantly exceeds performance of LoRA and DoRA with a small proportion of trainable parameters(5\%<).
For layers without selected sub-matrices, SMT freezes these layers, saving all their backward propagation computational cost, parameters update computational cost, optimizer memory cost, and activation memory cost.
For layers with selected sub-matrices, SMT reduces the computational costs of backward propagation and parameter updates, as well as the optimizer and activation memory costs, to less than 1\% of those incurred by standard full tuning (FT).


Below are our key contributions:
\begin{itemize}
    \item \textbf{PEFT SOTA Algorithm}
        We propose a novel fine-tuning method (SMT) which achieves state-of-the-art performance in parameter-efficient fine-tuning, effectively closing the gap between SMT and full fine-tuning.
        In contrast, LoRA and DoRA's performance saturation is faster than that of SMT as the number of trainable parameters grows. 
    \item \textbf{Language Model Anatomy:}
        We investigate the distinct impacts of attention mechanisms versus MLPs (Multi-Layer Perceptrons) in LLMs. Our findings indicate that attention layers are more critical than MLPs for downstream performance.
        Among the Q,K,V vectors of attention layers, we found V to be the most influential for performance.
    \item \textbf{Large Language Model System Efficiency:}
        The SMT implementation significantly reduces the computational cost of backward propagation, parameter updates, optimizer memory, and activation memory during fine-tuning.
        Our implementation is open source.

\end{itemize}


\section{Background and Related Works}
\label{sec:related_work}
Many works on parameter-efficient fine-tuning (PEFT) \cite{peft} have aimed to improve efficiency and performance by only fine-tuning lower-dimensional parameterization of model weights. 
Notable examples include LoRA \cite{hu2021lora}, DoRA \cite{liu2024dora}, QLoRA \cite{dettmers2023qlora}, and several other variants \cite{liu2024alora, dettmers2023qlora}. 
However, the results of these works still indicate a performance gap between PEFT methods and full fine-tuning (FT).
Concurrent research \cite{biderman2024lora} empirically demonstrated that such a gap is difficult if not impossible to eliminate, they also notice the performance saturation issue of LoRA, as we will discuss in Section \S\ref{sec:Experiments_Plateau}.

\begin{figure}[h]
    \centering
    \includegraphics[width=147mm]{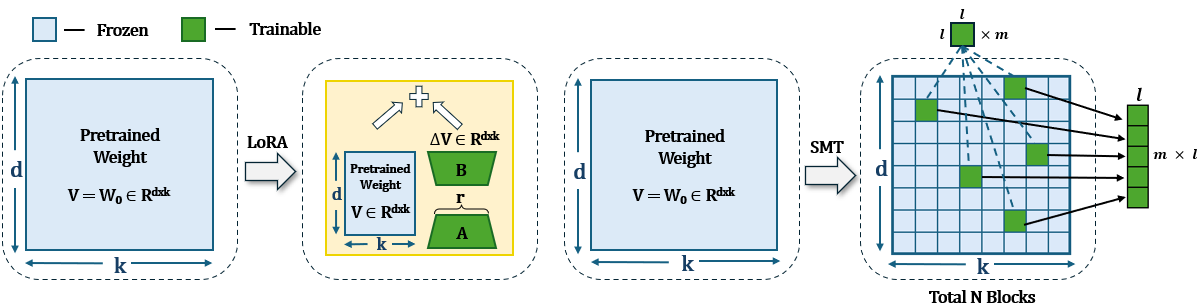}
    \centering
    \caption{Differences between low-rank adaption method LoRA and SMT. Upper picture dedicates adaption approach in LoRA and lower picture represents the sub-matrices sparsity approach in SMT.}
    \label{fig:SMT_vs_Apdaptors}
\end{figure}

Beside low-rank adaptation methods, sparsity-inspired approach is a natural alternative to reduce computational costs and memory footprint. As they have been recently applied to accelerate LLM inference, H$_2$O \cite{zhang2024h2o} leveraged sparsity in KV cache eviction policy; DeepSparse \cite{kurtic2023sparse} used L2-based distillation to promote sparsity.
In the case of sparsity-inspired fine-tuning, \cite{song2023sparse} developed a Sparse Increment Fine-Tuning (SIFT) approach to reduce GPU memory cost. However, SIFT \cite{song2023sparse} still requires full backward propagation to calculate all gradients, having no speed advantages compared to full fine-tuning(FT). 
Moreover, SIFT maps discontinuous memory gradients to continuous memory addresses, creating a significant time bottleneck and resulting in longer fine-tuning time than FT.
Our work builds upon existing strategies, but unlike previous approaches, our matrix sparsity method directly incorporates task-specific gradient information to dynamically adjust the sparsity level for optimization, and we achieve speedup as we describe in Section~\S\ref{sec:mem_comp_smt_vs_LoRA}. 

Suppose we are given a pre-trained auto-regressive language model $P_\Phi (y|x)$ parameterized by $\Phi$. Each downstream task is represented by a training dataset of context-target pairs: $Z = {(x_i, y_i)}_{i=1,\dots,N}$, where both $x_i$ and $y_i$ are sequences of tokens.
Equation (\ref{eq:lora}) dedicates the LoRA \cite{hu2021lora} fine-tuning process to maximize the conditional language modeling objective, which uses a low-rank representation to encode task-specific parameters.  
Specifically, LoRA freezes the pre-trained model weights and injects trainable rank decomposition matrices into each layer of the Transformer architecture. 
This is formulated as $\Delta \Phi = \Delta \Phi(\Theta)$, where $\Theta$ represents a much smaller-sized set of parameters with $|\Theta| \ll |\Phi_0|$. The resulting increment $\Delta \Phi$ can be as small as 0.01\% of the pre-trained weights parameter size $|\Phi_0|$ in gradient updates. This greatly reduces the number of trainable parameters and the GPU memory requirement while maintaining or even enhancing model performance.
\small
\begin{equation}\label{eq:lora}
\max_{\Theta} \sum_{(x,y) \in \mathcal{Z}} \sum_{t=1}^{|y|} \log (P_{\Phi_0+\Delta\Phi(\Theta)}(y_t | x, y_{<t}))
\end{equation}
\normalsize
In our work, our proposed Sparse Matrix Tuning(SMT) uses matrix sparsity as the parameter-efficient approach. In SMT's case, reusing Equation(~\ref{eq:lora}), the $\Theta$ represents the sub-matrices within the sparse weight matrices. SMT only fine-tunes sparse sub-matrices $\Theta$ instead of fine-tuning the whole pre-trained weight. Figure~\ref{fig:SMT_vs_Apdaptors} illustrates the differences between weight low-rank adaption method LoRA and our proposed sparse matrices tuning approach SMT. For a pre-trained weight matrix $W_0$, LoRA constrains its update by representing the latter with a low-rank decomposition $W_0 + \Delta W = W_0 + BA$, where $B\in R^{d\times r}$, $A\in R^{d\times r}$, and the rank $r \ll min(d,k)$. In SMT, we slice the pre-trained weight into $N$ sub-matrices and only fine-tune selected $M$ sub-matrices. The dimension of sub-matrices is $l \times l$, the total number of sub-matrices $N$ in a pre-trained weight is $N = \frac{d \times k }{l \times l}$.
SMT constrains its update by representing the latter with a sparse gradient matrix $\Delta W_{M}$,  $W_0 + \Delta W = W_0 + \Delta W_{M}$, where the number of fine-tuning sub-matrices $m \ll N$. 

Since our proposed SMT method focuses on fine-tuning sub-matrices which are most relevant to downstream tasks' performance, identifying these sub-matrices is non-trivial.
Previous works \cite{zhu2020modifying}, MEMIT \cite{meng2022mass}, and  \cite{geva2020transformer,geva2022transformer} indicated that feed-forward MLP layers of the LLMs are most influential. However, through our experiment analysis\S~\ref{sec:att-vs-mlp}, we found attention layers to be more relevant for performance than MLP layers.

\section{Methodology}
\label{sec:methodology}
\subsection{Selecting the most influential sparse sub-matrices}

\begin{figure}[htbp]
    \centering
    \begin{minipage}{0.55\textwidth}
        \centering
        \includegraphics[width=\textwidth]{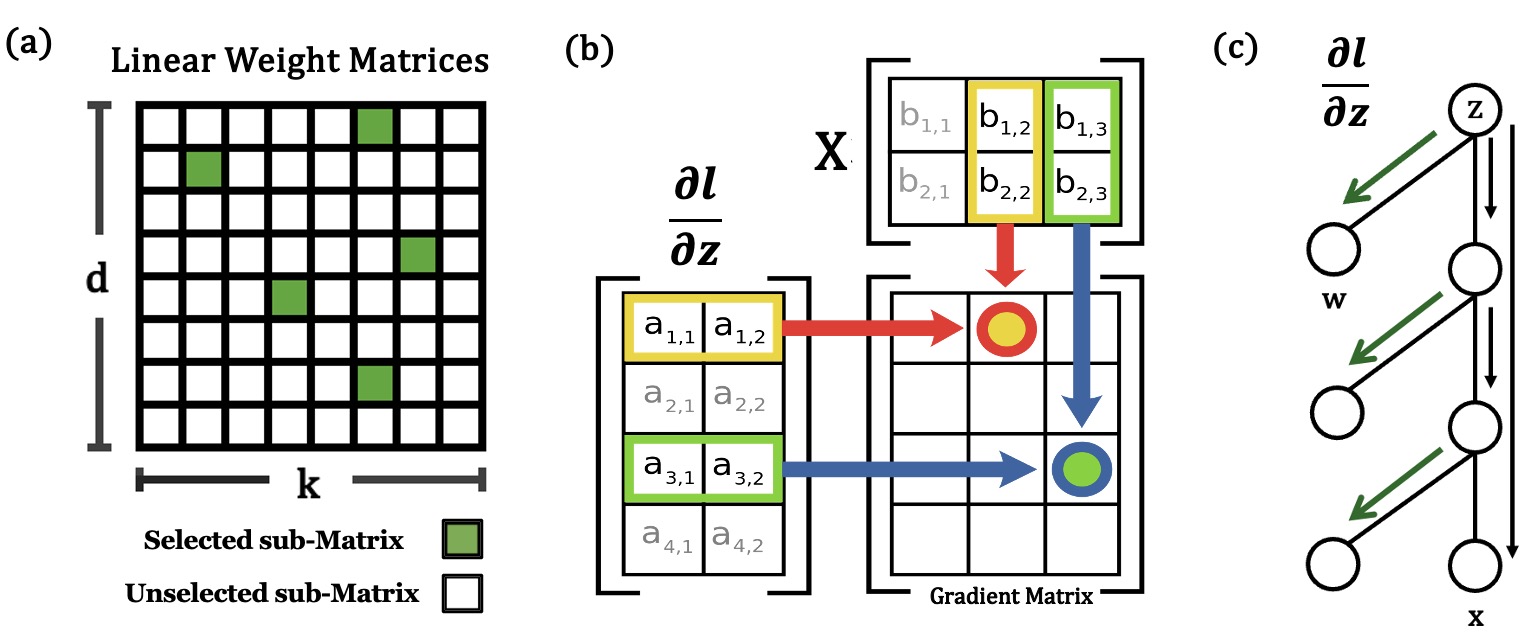}
        \caption{(a) A sparse weight matrix \textbf{$W$}. The green sub-matrices with significant gradients can be updated. (b) Backward propagation calculation for partial gradient for weight matrix $w$. (c) Computation graph in auto-differential systems.}
        \label{fig:sparse_matrix_system_implement}
    \end{minipage}
    \hfill
    \begin{minipage}{0.40\textwidth}
        \centering
        \captionof{table}{The experiments involved Full Fine-Tuning, SMT, LoRA, and DoRA on 4$\times$ A100 40GB GPUs using data parallel, with a batch size of 16. Communication between the GPU and CPU was facilitated via PCIe-G4.}
        \label{tab:time-profiling-results}
        \adjustbox{max width=\textwidth}{
            \begin{tabular}{p{2.4cm} c c c c c c c c c c c}
                \toprule
                \multicolumn{4}{c}{\centering\textbf{LLaMA-7B}} \\
                \midrule
                 \centering\textbf{PEFT method} & \textbf{\#Params\%} & \textbf{Time/s} & \textbf{Speedup}  \\
                \midrule
                \centering\textbf{Full Fine-tuning} & 100 & 243.84 & 1$\times$  \\
                \centering\textbf{SMT}              & 1.26 & 16.68 & 14.6$\times$  \\
                \centering\textbf{LoRA}             & 1.26 & 17.82 & 13.6$\times$  \\
                \centering\textbf{DoRA}             & 1.27 & 18.04 & 13.5$\times$  \\
                \bottomrule
            \end{tabular}
        }
    \end{minipage}
\end{figure}

Our methodology centers around the application of a \emph{\textbf{matrix sparsity}} approach during fine-tuning LLMs.
Specifically, we select certain sub-matrices in weight matrices within the model's weight matrices that exhibit maximal gradient changes during a 100 iterations warm-up phase (\cref{fig:sparse_matrix_system_implement}.a) at the beginning of fine-tuning.
Intuitively, our approach aims to select and modify the sub-matrices most relevant to the fine-tuning sub-task. 
When fine-tuning LLMs, SMT improves computational efficiency and reduces memory needs during fine-tuning by not updating a portion of weight matrices after the warm-up phase and storing sparse weights in compressed form. 

First and for most, SMT reduces \textbf{\textit{the backward computation costs with respect to weight}}  to 0.5\% of those associated with Full Fine-tuning (FT). 
SMT achieves this by decreasing the computation cost of gradients during backward propagation, as gradients are calculated for only a subset of the weights.
For linear layers in LLMs, where $Z = Wx$, the gradients with respect to weight matrix W and input x can be calculated as Equation (\ref{eq:backward_gradient}):

\begin{equation}\label{eq:backward_gradient}
\nabla_x f(x) = \frac{\partial l}{\partial Z} \cdot W; \hspace{7mm}
\nabla_W f(x) = \frac{\partial l}{\partial Z} \cdot x
\end{equation}

where $\partial l/\partial z$ is the gradient information from backward propagation in (\cref{fig:sparse_matrix_system_implement}.b,c). $\nabla_wf(x)$ is the gradient matrix and and $x$ is the activation in the (\cref{fig:sparse_matrix_system_implement}.b). (\cref{fig:sparse_matrix_system_implement}.b) also illustrates that only partial backward computations are necessary when we update selected sparse matrices. 
To calculate the sub-matrix gradient (highlighted in yellow), it is only necessary to multiply the yellow row in $\partial l/\partial z$ with the yellow column in the activation $x$.
Similarly, to calculate the green sub-matrix gradient, we only need to multiply the green row in $\partial l/\partial z$ with the green column in activation $x$.
Note that in backward propagation, we can only reduce computation when derivative to gradient matrix $w$ among as illustrated by the green arrows in (\cref{fig:sparse_matrix_system_implement}.c). but not other necessary computation. (black arrows) 

Besides, SMT reduces \textbf{\textit{the activation memory cost for the in forward pass}} to 0.5\%. Since SMT only computes the partial gradient, it only saves the relevant portions of activation $X$ necessary for the gradient calculation as is represented in Equation.~\ref{eq:backward_gradient}.
In (\cref{fig:sparse_matrix_system_implement}.b), to calculate the green and yellow gradients in the gradient matrix, we only need to save the yellow and green columns of the activation $X$.
This approach reduces the memory cost for the forward pass of the selected linear layer.

In addition, SMT reduces \textbf{\textit{the memory cost of the optimizer gradients}} to 0.5\%. Since SMT only updates selected sparse sub-matrices, only partial gradients are stored. This approach significantly cuts the memory cost of the Adam optimizer to 0.5\%. This reduction is crucial because the memory cost of the Adam optimizer is typically twice the size of the model, which often consumes the majority of GPU RAM.
 
Furthermore, SMT reduces \textit{\textbf{the gradient step computation cost}} to 0.5\%. By updating selected sparse sub-matrices, SMT performs partial gradient steps and significantly reduce step computation cost. 

In SMT, all the layers except selected Q, K, and V vectors are \textbf{\textit{frozen}} during fine-tuning. By doing this, SMT avoids all the weight backward propagation computational cost, parameters update computational cost, optimizer memory cost, and activation memory cost in frozen layers. The rationale for fine-tuning only the Q, K, and V vector is detailed in Section~\S{\ref{sec:att-vs-mlp}}.

By applying sparse sub-matrix fine-tuning, SMT can reduce the fine-tuning memory cost of LLaMA-7B and LLaMA2-7B to less than 20GB and fit the fine-tuning into a 3090 24GB GPU. We also reduce the computation and achieve faster fine-tuning compared with FT and LoRA/DoRA, Section \S~\ref{sec:mem_comp_smt_vs_LoRA} provides more details.

\subsection{Implementation} 
In SMT, we first \textbf{\textit{sum up gradient}} from the attention linear layers in every single warm-up iterations. The sum up gradient information are used to identify task-specific sparse blocks. After the warm-up steps, we average the absolute values within the sub-matrices, select the sub-matrices with the largest value, and save the indices for the selected sub-matrices. In all of our experiments, we use $l \times l= 256 \times 256$ as sub-matrices block size. During the warm up steps, we can apply offload \cite{rajbhandari2020zero} on memory constraint GPU devices. 
Since SMT requires fewer than 100 warm-up steps in our experiments, it does not become a bottleneck during fine-tuning epochs.
Additionally, SMT implements \textbf{\textit{a custom sparse linear layer}} to ensure that unselected gradients are not calculated, saved, and updated (Code Snippet~\ref{fig:code_sparse_linear_layer}). We replace the selected linear layers with these customized sparse linear layers.

The custom sparse linear layer applies \textbf{\textit{a specialized sparse linear multiplication function}}, integrated into our customized sparse linear layers (Code Snippet~\ref{fig:code_forward}). This function calculates partial weight gradients based on the input, weight, and selected weight index. It significantly reduces the computational cost of backward propagation weight gradients to just 0.5\% and minimizes the memory usage of the returned partial gradients to only 0.5\%.

The specialized sparse linear multiplication function \textit{\textbf{rewrites both forward and backward functions.}} In the forward pass (Code Snippet~\ref{fig:code_forward}) of sparse linear multiplication function, we only save selected activation $x$ using \texttt{ctx.save\_for\_backward()}, and in the backward pass (Code Snippet~\ref{fig:code_backward}), we customize matrix multiplication to calculate the needed partial gradients given partial input and gradient index(shown in Figure \ref{fig:sparse_matrix_system_implement}(b)).
It is important to note that we do not use Sparse Matrix-Matrix Multiplication(SPMM)\footnote{Sparse Matrix-Matrix Multiplication (SPMM) is significantly slower than General Matrix Multiply (GeMM)} 
because we concatenate the selected sparse sub-matrices and formed a $m \times l \times l$ dense matrix as illustrated in right part of figure 1. This would not cost additional time since memory allocations remain continuous within each sub-matrix. Despite employing matrix sparsity, we still leverage the advantages of dense matrix multiplication.

Furthermore, SMT gathers sparse matrix but still leverages dense matrix. SMT customizes the function for\textbf{\textit{ gathering trainable parameters.}} This customized function selectively gathers weight sub-matrices in the Q, K, and V vector layers and passes them to the Adam optimizer. By continuing to use the well-designed \texttt{FusedAdam} from the \texttt{deepspeed} library \cite{aminabadi2022deepspeed}, we maintain the computational speed of dense matrix weight updates. 
However, our approach reduces the gradient memory cost in the optimizer to just 0.5\%.
\subsection{Memory and Computation Saving: SMT vs. Low-rank Adaption Methods}
\label{sec:mem_comp_smt_vs_LoRA}
\textbf{\textit{SMT is more computational efficient than weight low-rank adaption method}} when the number of trainable parameters are the same, weight low rank adaption methods need to maintain additional adapters, which require additional forward computation. For instance, since LoRA maintains adaptors $A$ and $B$, and the forward propagation is:
\begin{equation}\label{eq:gradient}
h = W_0x + \Delta W_x = W_0x + BAx
\end{equation}
where the term $BAx$ requires additional forward propagation calculation, which is cut off in SMT. Regarding memory cost, since SMT does not require additional low-rank adapters  $A$ and $B$, SMT can achieve \textbf{\textit{lower memory cost}} than LoRA and DoRA under the same amount of trainable parameters setting. We illustrate this in Figure~\ref{fig:SMT_vs_Apdaptors}, by abandoning additional adaptor weight $A$ and $B$, SMT can achieve lower memory cost.
Taking the popular LLaMA-13B model as an example, since the model size is approximately 25 GB, if we fine-tune 1\% of parameters, SMT can potentially save 250MB GPU meory.
In Table~\ref{tab:time-profiling-results}, we provide the fine-tuning time costs for SMT, Full Fine-tuning, LoRA, and DoRA. SMT achieves an 14.6× speedup compared to Full Fine-tuning and outperforms both LoRA and DoRA.
We conducted time profiling by averaging the fine-tuning time every 10 iterations over 1000 iterations, following a 500-iteration warm-up period. Full fine-tuning utilized offload settings to accommodate the LLaMA model, which employs the Adam optimizer, within the 40GB GPU.

\section{Experiments and Results}
\label{sec:Experiments}
\subsection{Experimental Settings}
\label{sec:Experiments_setting}

\textbf{Model Architecture and Dataset:} In our experimental setup, we use open-weight LLaMA-7B, LLaMA-13B, LLaMA2-7B, and LLaMA3-8B models \cite{llama3modelcard}. In Subsection\S~\ref{sec:Experiments_commonsense}\S~\ref{sec:Experiments_Plateau}, We perform fine-tuning on the Common Sense Reasoning tasks with 8 sub-tasks, each with a predefined training and testing set. We follow the setting of \cite{hu2023llm, liu2024dora} and amalgamate the training datasets from all 8 tasks to create the final training dataset \texttt{commonsense\_170k} and conduct evaluations on the individual testing dataset for each task. We calculate an average score to encapsulate the overall efficacy. In Subsection\S~\ref{sec:Experiments_math}, we perform fine-tuning on \texttt{Math10K} \cite{hu2023llm} dataset which includes \texttt{MultiArith}, \texttt{GSM\_8K} \cite{cobbe2021gsm8k}, \texttt{AddSub}, 
\texttt{AQuA}, \texttt{SingleEq}, \texttt{SVAMP} datasets and evaluate the efficiency on their testsets.

\textbf{Training Framework and SMT Hyper-parameters:} We used the \texttt{DeepSpeed} \cite{aminabadi2022deepspeed} library for fine-tuning and \texttt{accelerate} \cite{accelerate} library for inference evaluation. Both training and fine-tuning are using \texttt{dtype} bf16. All experiments are fine-tuned for 3 epochs.
In all our experiments in Section\S~\ref{sec:Experiments}, sub-matrices are selected in blocks of size $l = 256$. 
We choose this specific sub-matrix dimension $l$ because it is the largest common factor of the column and row sizes of all linear layers in the LLaMA series models, using this dimension for slicing avoids remainder issues. We freeze all MLP layers and apply SMT only to the Q, K, and V vectors in the attention mechanism. 
In Section\S\ref{sec:att-vs-mlp}, we explain the rationale why we only apply SMT to attention mechanism instead of MLP.
At the end of the gradient warm-up iteration, SMT ranks the average absolute gradient values within each sub-matrix and selects those with the highest average values. The rationale of such selection is explained more in detail in Appdendix~\ref{appendix:sub_matrices_selection}.
We apply 100 warm-up iterations to all SMT experiments on \texttt{Commonsense} dataset and apply 25 warm-up iterations to all SMT experiments on \texttt{Math10K} dataset.

\textbf{PEFT Baselines:} For state-of-the-art (SOTA) baselines, we choose to include LoRA \cite{hu2021lora} and DoRA \cite{liu2024dora}, both of them focus on fine-tuning using low-rank adaptation method. 

\textbf{Computational Resources:} We conduct our experiments and SOTA baselines of LoRA \cite{lora} and DoRA \cite{dora} to fine-tune LLaMA-7B and LLaMA2-7B model with 4 NVIDIA A100\_40GB GPUs and fine-tune LLaMA-13B and LLaMA3-8B model with 4 NVIDIA A100\_80GB GPUs. Communication between the CPU and GPU was facilitated via PCIe-G4 and communication between GPUs was facilitated via Nvlink-3.

\textbf{Evaluation Metrics:}
We have evaluated the performance of SMT in terms of computational efficiency (wall-clock time speedup), memory usage (analysis for memory complexity) in methodology Section\S~\ref{sec:methodology}. In this section, we will mainly evaluate SMT in terms of popular NLP tasks to test its ability to generalize to all downstream tasks. In Subsection\S~\ref{sec:Experiments_commonsense}\S~\ref{sec:Experiments_Plateau}, we evaluate the performance of SMT on 8 tasks in the \texttt{Commonsense} dataset, including \texttt{BoolQ}, \texttt{PIQA}, \texttt{SIQA}, \texttt{HellaSwag}, \texttt{ARC-e}, \texttt{ARC-c}, and \texttt{OBQA}, and we calculate an average score to encapsulate the overall efficacy.In Subsection\S~\ref{sec:Experiments_math}, we perform fine-tuning on \texttt{Math10K} \cite{hu2023llm} dataset which includes \texttt{MultiArith}, \texttt{GSM\_8K}, \texttt{AddSub}, 
\texttt{AQuA}, \texttt{SingleEq}, \texttt{SVAMP} datasets and evaluate the efficiency of SMT on their testsets. All of the experiments are evaluated using accuracy.

\subsection{Commonsense Reasoning}
\label{sec:Experiments_commonsense}
We evaluate SMT against the state-of-the-art(SoTA) weight low-rank adaptor method includes LoRA and DoRA. To ensure 
a fair comparison, we fine-tuned model with SMT following the LoRA and DoRA configuration.
We ensure all the hyper-parameters including batch size, data type, learning rate, and sequence length are identical to what was reported in LoRA and DoRA \cite{hu2021lora, liu2024dora}.
We re-implemented LoRA and DoRA and achieved their best performance reported in  \cite{liu2024dora}. 

Table~\ref{tab:commonsense_results} demonstrates that SMT consistently surpasses baseline methods across LLaMA-7B, LLaMA13B, LLaMA2-7B, and LLaMA3-8B. Notably, by overcome plateau phenomenon, SMT further enhances accuracy of DoRA by 3.0\%, 2.8\%, 2.9\%, and 2\% on LLaMA-7B, LLaMA-13B, LLaMA2-7B, and LLaMA3-8B respectively.
Notably, LoRA and DoRA will not achieve better performance with larger trainable parameters and exhibit the plateau phenomenon.
In Subsection\S~\ref{sec:Experiments_Plateau}, we report and demonstrate the plateau issue in LoRA and DoRA and demonstrate SMT overcomes this issue.
Moreover, by fine-tuning less than 5\% of all parameters, SMT achieves similar accuracy performance of full fine-tuning while speedup 14.6$\times$ (speedup details in Table ~\ref{tab:time-profiling-results}) and save 99.5\% of optimizer memory(memory bottleneck in fine-tuning, details discussed in Section\S~\ref{sec:methodology}). 

SMT can also consistently surpass LoRA and DoRA under the same number of trainable parameters where LoRA and DoRA achieve the best results, SMT can surpass their performance and also outstrip
ChatGPT-3.5-turbo\footnote{Results of ChatGPT-3.5-turbo  are reported in DoRA \cite{dora}}. 
For instance, SMT consistently
surpasses DoRA on LLaMA2-7B, LLaMA3-8B, LLaMA-13B, and LLaMA-7B by 1.3\%, 1.6\%, 0.9\%, and 0.3\% respectively, under their best performance trainable parameter number.

\begin{table*}[t!]
\centering
\caption{Accuracy comparison of LLaMA 7B, LLaMA 13B, LLaMA2 7B, and LLaMA3 8B with various PEFT methods on eight commonsense
reasoning datasets. Results of all the baseline methods on LLaMA 7B, LLaMA 13B, LLaMA2 7B, LLaMA3 8B are taken from \cite{liu2024dora}. Results of all SMT are obtained using the hyper-parameters described in \cite{liu2024dora} under the same settings. Bold texts dedicate the performance of SMT under the same numbers of parameters where LoRA and DoRA achieve the best performance. Blue texts dedicate the best performance of SMT. Please note that the performance of LoRA and DoRA\textbf{\textit{ under larger numbers of trainable parameters can be found in Table~\ref{tab:plateau}.}}} 

\label{tab:commonsense_results}
\resizebox{\textwidth}{!}{
\setlength{\tabcolsep}{1.5mm}{
\begin{tabular}{p{1.5cm} c c c c c c c c c c c}
\toprule
\centering\textbf{Model} & \textbf{PEFT method} & \textbf{\#Params\%} & \textbf{BoolQ} & \textbf{PIQA} & \textbf{SIQA} & \textbf{HellaSwag} & \textbf{WinoGrande} & \textbf{ARC-e} & \textbf{ARC-c} & \textbf{OBQA} & \textbf{AVG} \\
\midrule
\textbf{ChatGPT(175B)} & \textbf{-} & - & 73.1 & 85.4 & 68.5 & 78.5 & 66.1 & 89.8 & 79.9 & 74.8 & 77.0 \\
\midrule
\multirow{4}{*}{\textbf{LLaMA-7B}}  
& \textbf{LoRA(Best)} & 0.83 & 67.5 & 80.8 & 78.2 & 83.4 & 80.4 & 78.0 & 62.6 & 79.1 & 76.3 \\
& \textbf{DoRA(Best)} & 0.84 & 69.7 & 83.4 & 78.6 & 87.2 & 81.0 & 81.9 & 66.2 & 79.2 & 78.4 \\
& \textbf{SMT}        & 0.84 & 68.7 & 81.7 & 78.3 & 91.6 & 78.8 & 84.1 & 68.7 & 77.4 & \textbf{78.7} \\
& \textbf{SMT(Best)}  & 4.91 & 72.0 & 82.9 & 80.7 & 93.3 & 82.4 & 86.1 & 70.6 & 83.0 & \textbf{\textcolor{blue}{81.4}} \\
& \textbf{Full Fine-tuning}  & 100 & 69.9 & 84.2 & 78.9 & 92.3 & 83.3 & 86.6 & 72.8 & 83.4 & 81.4 \\

\midrule
\multirow{4}{*}{\textbf{LLaMA-13B}}  
& \textbf{LoRA(Best)} & 0.67 & 72.1 & 83.5 & 80.5 & 90.5 & 83.7 & 82.8 & 68.3 & 82.4 & 80.5 \\
& \textbf{DoRA(Best)} & 0.68 & 72.4 & 84.9 & 81.5 & 92.4 & 84.2 & 84.2 & 69.6 & 82.8 & 81.5 \\
& \textbf{SMT}  & 0.68 & 71.1 & 84.4 & 81.7 & 93.7 & 83.2 & 86.7 & 73.7 & 85.2 & \textbf{82.4} \\
& \textbf{SMT(Best)}  & 4.91 & 72.6 & 86.1 & 81.9 & 95.0 & 86.1 & 88.2 & 77.1 & 87.4 & \textbf{\textcolor{blue}{84.3}} \\

\midrule
\multirow{4}{*}{\textbf{LLaMA2-7B}}  
& \textbf{LoRA(Best)} & 0.83 & 69.8 & 79.9 & 79.5 & 83.6 & 82.6 & 79.8 & 64.7 & 81.0 & 77.6 \\
& \textbf{DoRA(Best)} & 0.42 & 72.0 & 83.1 & 79.9 & 89.1 & 83.0 & 84.5 & 71.0 & 81.2 & 80.5 \\
& \textbf{SMT}  & 0.84 & 72.0 & 83.8 & 80.8 & 93.3 & 82.8 & 86.7 & 74.0 & 81.0 & \textbf{81.8} \\
& \textbf{SMT(Best)}  & 4.91 & 72.6 & 85.2 & 82.0 & 94.4 & 85.7 & 87.8 & 74.5 & 85.0 & \textbf{\textcolor{blue}{83.4}} \\
& \textbf{Full Fine-tuning}  & 100 & 72.8 & 83.4 & 78.7 & 92.7 & 85.5 & 86.2 & 74.7 & 83.4 & 82.2\\

\midrule
\multirow{4}{*}{\textbf{LLaMA3-8B}}  
& \textbf{LoRA(Best)} & 0.70 & 70.8 & 85.2 & 79.9 & 91.7 & 84.3 & 84.2 & 71.2 & 79.0 & 80.8 \\
& \textbf{DoRA(Best)} & 0.71 & 74.6 & 89.3 & 79.9 & 95.5 & 85.6 & 90.5 & 80.4 & 85.8 & 85.2 \\
& \textbf{SMT}        & 0.71 & 75.7 & 88.4 & 81.4 & 96.2 & 88.2 & 92.7 & 83.2 & 88.6 & \textbf{86.8} \\
& \textbf{SMT(Best))}& 3.01 & 75.1 & 89.9 & 82.4 & 96.3 & 88.8 & 92.6 & 82.8 & 89.6 & \textcolor{blue}{\textbf{87.2}} \\

\bottomrule
\end{tabular}
}}
\end{table*}

\subsection{Plateau in Weight low rank adaption methods}
\label{sec:Experiments_Plateau}
\begin{figure}[h]
    \centering
    \includegraphics[width=128mm]{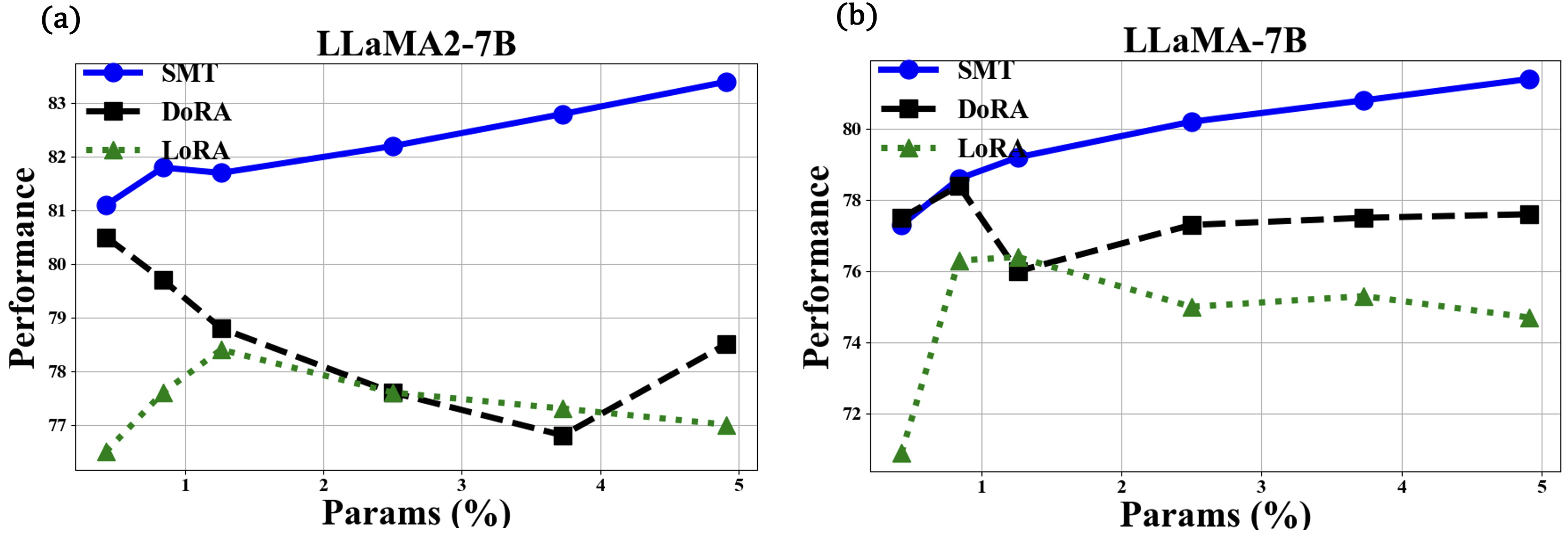}
    \centering
    \caption{Accuracy comparison of LoRA, DoRA, and SMT under different scaling of trainable parameters on commonsense reasoning datasets.}
    \label{fig:plateau_}
\end{figure}

In Table~\ref{tab:plateau} and Figure~\ref{fig:plateau_}, we scale up the model size and presents how the performance of LoRA and DoRA will be under larger number of trainable parameters. We reimplement all the experiments of LoRA \cite{lora} and DoRA \cite{dora} using their official repository and followed their recommendation of hyper-parameters to achieve best performance under every single trainable parameter size. 
We observe that both DoRA and LoRA models, with some larger ranks, their performance slightly degrades.
However, SMT continues improving its performance when we scale up the trainable parameter size. When we scale up the trainable parameter size to 4.91\%, SMT significantly surpass DoRA by 3.8\% and 4.9\% on LLaMA-7B and LLaMA-2-7B fine-tuned models.
We postulate that such plateau phenomenon of LoRA or DoRA is due to their lossy low-rank approximation of the full weight information (includes lots of noise), whereas our SMT focuses on most prominent submatrices (contains less noise) and remains full rank gradient updates for the selected portion, making SMT performs better.

\begin{figure}[H]
    \centering
    \begin{minipage}{0.60\textwidth}

        \centering
        \captionof{table}{Accuracy comparison of LoRA, DoRA, and SMT under different scaling of trainable parameters on commonsense reasoning datasets. Given certain base model and PEFT method, we gradually increase the number of trainable parameters on each line from left to right. On each line, the best performing model has \textcolor{blue}{$^*$}. }
        \label{tab:plateau}
        \adjustbox{max width=\textwidth}{

            \begin{tabular}{p{1.55cm} c c c c c c c c c c c}
            \toprule
            \textbf{} & \textbf{Method} & \textbf{0.43} & \textbf{0.84} & \textbf{1.26} & \textbf{2.50} & \textbf{3.73} & \textbf{4.91} \\
            \midrule
            
            \centering\multirow{3}{*}{\textbf{LLaMA-7B}}  
            & \textbf{LoRA} & 70.9 & \textbf{76.3}\textcolor{blue}{$^*$} & 76.4 & 75.0 & 75.3 & 74.7 \\
            & \textbf{DoRA} & 77.5 & \textbf{78.4}\textcolor{blue}{$^*$} & 76.0 & 77.3 & 77.5 & 77.6 \\
            & \textbf{SMT}  & 77.3 & 78.6 & 79.2 & 80.2 & 80.8 & \textbf{81.4}\textcolor{blue}{$^*$} \\
            \midrule
            
            \multirow{3}{*}{\textbf{LLaMA2-7B}}  
            & \textbf{LoRA} & 76.5    & \textbf{77.6}\textcolor{blue}{$^*$} & 78.4 & 77.6 & 77.3 & 77.0 \\
            & \textbf{DoRA} & \textbf{80.5}\textcolor{blue}{$^*$} & 79.7 & 78.8 & 77.6 & 76.8 & 78.5 \\
            & \textbf{SMT}  & 81.1 & 81.8 & 81.7 & 82.2 & 82.8 & \textbf{83.4}\textcolor{blue}{$^*$} \\
            
            \bottomrule
            \end{tabular}

        }
    \end{minipage}
    \hfill
    \begin{minipage}{0.38\textwidth}
        
        \centering
        \captionof{table}{ Fine-tuned LLaMA-7B model performance on \texttt{Commonsense}. AVG dedicates the average test score of eight subsets among \texttt{Commonsense}. MLP\% and Attention\% presents the percentage of trainable parameters apply to MLPs and attention mechanisms respectively.}
        \label{tab:mlp_vs_arrention}
        \adjustbox{max width=\textwidth}{
        
            \begin{tabular}{p{1.5cm} c c c c}
            \toprule
            \textbf{Model} & \textbf{MLP\%} & \textbf{Attention\%} & \textbf{ AVG}  \\
            \midrule
            \multirow{4}{*}{\makecell{\textbf{SMT(0.84\%)} \\ \textbf{LLaMA-7B}}} 
             &  0.84 & 0   & 76.7 \\
             & 0.42 & 0.42 & 77.3 \\
             & 0.21 & 0.63 & 77.8  \\
             & 0    & 0.84 & 78.7 \\
            \bottomrule
            \end{tabular}
            
        }
    \end{minipage}
\end{figure}

\subsection{Other Dataset}
\label{sec:Experiments_math}

To ensure our findings above are generalizable, we further examine the performance of SMT under arithmetic reasoning dataset, \texttt{Math10K} \cite{hu2023llm}. \texttt{Math10K} dataset has six subsets including \texttt{GSM8k}, \texttt{SingleEq}, \texttt{SVAMP}, \texttt{MultiArith}, \texttt{AddSub}, and \texttt{AQuA}.
More details about \texttt{Math10K} dataset can be found in Appendix~\ref{appendix:math10k}. To ensure a fair comparison, we follow the open source hyper-parameter instruction in  \cite{hu2023llm} to achieve best performance for LoRA and Dora, and apply the same hyper-parameters to SMT while only fine-tune the learning rate. Table ~\ref{tab:mix_dataset_results} reports the performance of LoRA, DoRA, and SMT on the \texttt{Math10K} dataset. We can observe that SMT surpasses the best achievable performance of LoRA and DoRA by 1.3\% and 1.1\% respectively using the same amount of trainable parameters. In addition, by scaling up the trainable model size to 1.26\%, SMT achieves better performance and surpasses the best performance of LoRA and DoRA by 2.5\% and 2.3\% respectively. 

\begin{table*}[htbp]
\centering
\caption{SMT, LoRA and DoRA reproduction, and experiment results on \texttt{Math10K} dataset.}
\label{tab:mix_dataset_results}
\resizebox{\textwidth}{!}{
\setlength{\tabcolsep}{1.5mm}{
\begin{tabular}{p{1.5cm} c c c c c c c c c c c}
\toprule
\textbf{Model} & \textbf{PEFT method} & \textbf{\#Params\%} & \textbf{GSM8k} & \textbf{SingleEq} & \textbf{SVAMP} & \textbf{MultiArith} & \textbf{AddSub} & \textbf{AQuA} & \textbf{AVG} \\
\midrule
\multirow{3}{*}{\textbf{LLaMA-7B}}  
& \textbf{LoRA(Best)} & 0.86 & 35.4 & 83.2 & 52.1 & 92.8 & 83.4 & 18.6 & 60.9 \\
& \textbf{DoRA(Best)} & 0.86 & 35.2 & 83.7 & 51.8 & 92.8 & 82.8 & 20.2 & 61.1 \\
& \textbf{SMT}        & 0.86 & 34.2 & 84.6 & 53.6 & 91.5 & 85.8 & 23.6 & 62.2 \\
& \textbf{SMT(Best)}  & 1.26 & 35.6 & 85.3 & 54.8 & 93.4 & 86.8 & 24.2 & 63.4 \\

\bottomrule
\end{tabular}
}}
\end{table*}

\section{Further Dicussion}

\subsection{Attention versus MLP}
\label{sec:att-vs-mlp}
In order to study what components are more critical for LLM's downstream performance, we conduct ablation studies that compares MLPs vs. attention layers by adjusting the ratio of their trainable parameters respectively.
We apply SMT and fine-tune 0.86\% of parameters on LLaMA-7B using \texttt{Commonsense} dataset. In Table~\ref{tab:mlp_vs_arrention}, we present four experiments. In the first row, all trainable parameters are allocated to MLPs. In the second row, both MLPs and Q, K, V vectors from attention mechanisms receive 0.43\% of trainable parameters. In the third row, 0.62\% of trainable parameters are assigned to Q, K, V vectors from attention mechanisms and 0.21\% to MLPs. In the fourth row, all trainable parameters are dedicated to Q, K, V vectors from attention mechanisms. To guarantee a fair comparison, all the other hyper-parameters and settings are the same among these experiments.  

In these experiments, allocating X\% of trainable parameters to MLPs or attention mechanisms means ranking the average absolute gradient values of each sub-matrix within the MLPs or attention mechanisms and selecting those with the highest average values until the number of parameters reaches X\%. The results reveal a significant performance gap between the first and fourth rows. The more trainable parameters we allocate to attention mechanisms, the better the fine-tuned model performs. When all SMT trainable parameters are applied to attention mechanisms, the model outperforms the one where all parameters are allocated to MLPs by 2.0\%. 
Our empirical findings challenge previous assumptions \cite{zhu2020modifying,meng2022locating,geva2020transformer,geva2022transformer} that the memory sections of large language models are primarily located in feed-forward MLP layers.


\subsection{V Vector Versus Q, K Vector}
\label{sec:QKV}

\begin{table*}[t!]
\centering
\caption{K SMT, Q SMT, and V SMT assign all trainable parameters to only K, or only Q, or only V vectors respectively, and fine-tuned 0.86\% of the parameters on LLaMA-7B using the Commonsense dataset. QKV SMT assign all trainable parameters to QKV vectors and select sub-matrices automatically.}
\label{tab:experiments_qkv}
\resizebox{\textwidth}{!}{
\setlength{\tabcolsep}{1.5mm}{
\begin{tabular}{p{1.5cm} c c c c c c c c c c c}
\toprule
\centering\textbf{Model} & \textbf{Param location} & \textbf{\#Params\%} & \textbf{BoolQ} & \textbf{PIQA} & \textbf{SIQA} & \textbf{HellaSwag} & \textbf{WinoGrande} & \textbf{ARC-e} & \textbf{ARC-c} & \textbf{OBQA} & \textbf{AVG} \\
\midrule
\multirow{4}{*}{\textbf{LLaMA-7B}}  
& \textbf{K SMT}    & 0.84 & 65.5 & 79.1 & 76.2 & 88.3 & 73.2 & 80.3 & 60.8 & 68.0 & 73.9 \\
& \textbf{Q SMT}    & 0.84 & 65.7 & 79.3 & 75.5 & 88.2 & 72.5 & 80.1 & 59.6 & 72.5 & 75.3 \\
& \textbf{V SMT}    & 0.84 & 68.7 & 82.1 & 78.1 & 91.6 & 78.8 & 83.0 & 68.7 & 77.2 & 78.5 \\
& \textbf{QKV SMT}  & 0.84 & 68.7 & 81.7 & 78.3 & 91.6 & 78.8 & 84.1 & 68.7 & 77.4 & 78.7 \\

\bottomrule
\end{tabular}
}}
\end{table*}

\begin{figure}[htbp]
    \centering
    \begin{minipage}{0.74\textwidth}
        \centering
        \caption{ A visualization of trainable Q, K, V layers when fine-tuning 0.86\% trainable parameters on LLaMA-7B. LLaMA-7B has 32 layers of MLPs, each contains a Q vector, a K vector, and a V vector. White layers are frozen and green layers contain trainable parameters.}
        \includegraphics[width=\textwidth]{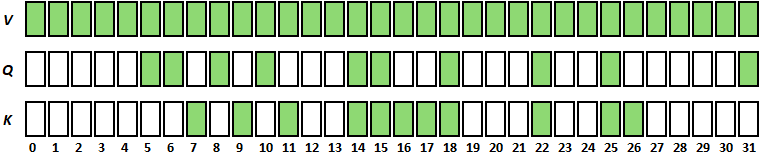}

        \label{fig:QKV_diff}
    \end{minipage}
    \hfill
    \begin{minipage}{0.24\textwidth}
        \centering
        \includegraphics[width=\textwidth]{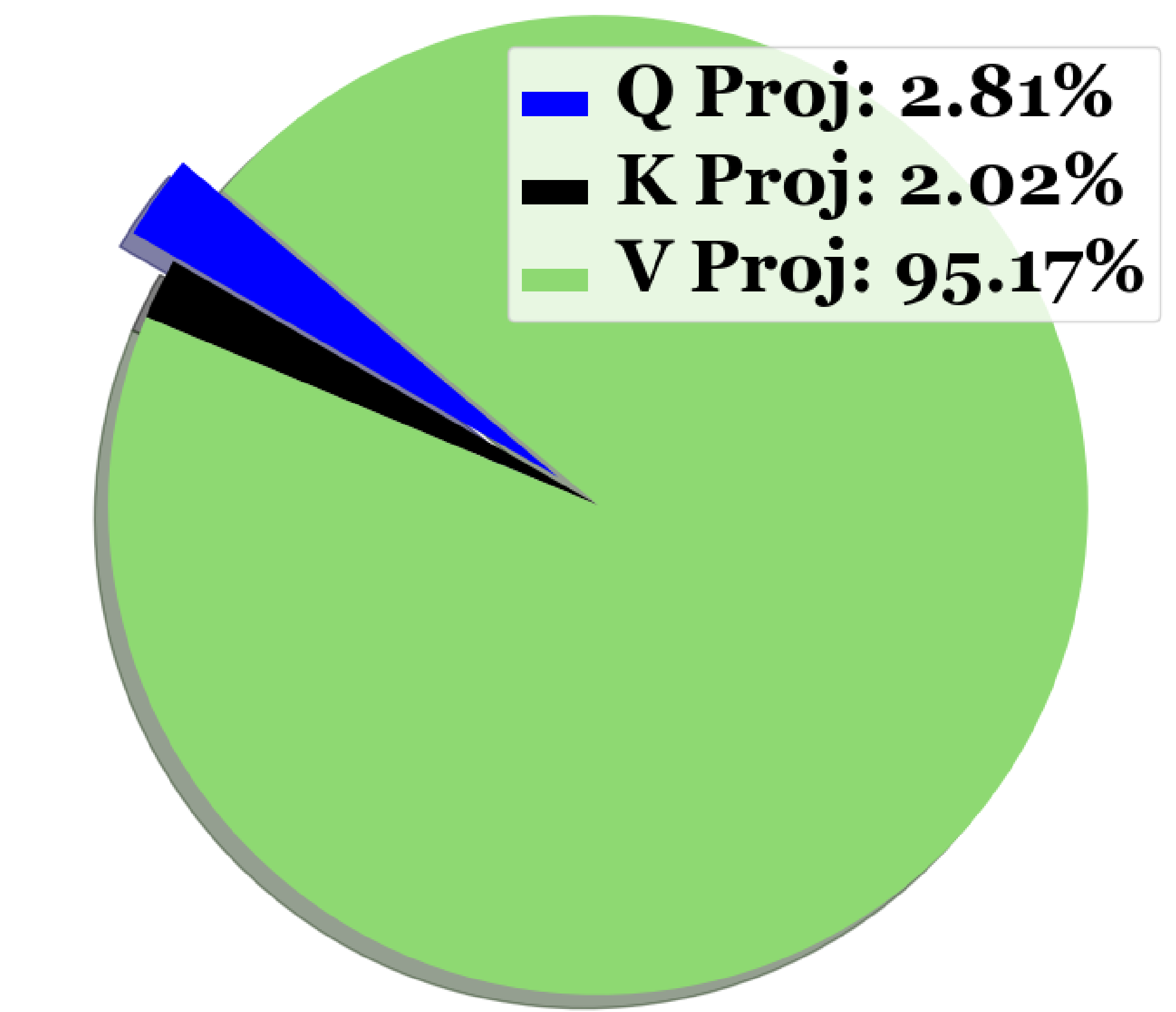}
        \vspace{-0.1in}
        \caption{Distribution of trainable parameters among Q, K, V.}
        \label{fig:qkv_pie_chat}
        
    \end{minipage}
\end{figure}

Based on our observation that attention is more critical, in all of our SMT experiments in Section\S~\ref{sec:Experiments}, we only allocate SMT trainable parameters to Q, K, V vectors from attention mechanisms. We rank the average absolute gradient values of every single sub-matrix within attention mechanisms and select those with the highest average values until the parameter ratio limit is reached.
Surprisingly, we observed that the trainable parameters are predominantly assigned to the V vectors.
In our ablation experiments, we experimented with assigning all trainable parameters to only K, or only Q, or only V vectors, and fine-tuned 0.86\% of the parameters on LLaMA-7B using the \texttt{Commonsense} dataset. 
As shown in Figure~\ref{fig:qkv_pie_chat}, 95.17\% of the trainable parameters are automatically assigned to the V vectors by SMT. Figure~\ref{fig:QKV_diff} indicates that all V vectors have trainable parameters, while 22 Q vectors and 21 K vectors are completely frozen.
This suggest that ~\textbf{\textit{the V vectors contain most of the memory}} and are the most significant among the Q, K, and V.

\vspace{-0.15 in}

Table~\ref{tab:experiments_qkv} presents four additional experiments where we fine-tuned 0.86\% of the parameters of LLaMA-7B using SMT on the \texttt{Commonsense} dataset.
In the first three rows, all trainable parameters are allocated to the K vectors, Q vectors, and V vector , respectively. In the fourth row, the trainable parameters are assigned to Q, K, V vectors directly and allocated by SMT automatically. The trainable parameters are distributed among the K, Q, and V vectors, as detailed in Figure~\ref{fig:qkv_pie_chat}, with the trainable states of the QKV layers shown in Figure~\ref{fig:QKV_diff}. 

The results show a significant performance gap when comparing the allocation of all trainable parameters to the V vectors versus the Q and K vectors. Assigning all parameters to the V vectors outperforms the K vectors by 4.6\% and the Q vectors by 3.2\%.
These observations suggest that the V vectors are the most significant among the Q, K, and V vectors; it also hints that SMT is able to effectively select sub-matrices containing crucial memory sections.



\section{Conclusion}
Our proposed Sparse Matrix Tuning (SMT) achieves SoTA performance, narrowing the gap between SMT and full fine-tuning. 
SMT can also reduce the computational cost of backward propagation, parameter updates, optimizer memory, and activation memory during fine-tuning to achieve 14.6$\times$ speedup.  
The empirical evidence presented in our extensive experiments suggests that attention layers are more critical than MLPs for downstream performance; V vector is the most influential vector for performance among Q, K, V vectors. 

\section{Acknowledgements}

We sincerely thank Sida Wang for her valuable contributions to this project, including developing and finalizing the implementation methodology, generating critical experimental results that strengthened our conference presentation, and supporting the open-source release of our codebase. \\
We also extend our gratitude to Qianou (Christina) Ma, Chenyang Yang, Chen Liu, and Yu (Ivy) Yang for the suggestion in paper writing and their valuable feedback. \\
This research used the Bridges-2 at Pittsburgh Supercomputing Center(PSC) and Delta advanced computing. Pittsburgh Supercomputing Center is supported by National Science Foundation grants \#2138259, \#2138286, \#2138307, \#2137603, and \#2138296. The Delta advanced computing is supported by the National Science Foundation (award OAC 2005572) and the State of Illinois. Delta is a joint effort of the University of Illinois Urbana-Champaign and its National Center for Supercomputing Applications.  \\
Though Heather Miller and Juncheng Billy Li are employees of Two Sigma Investments, this work was performed independently from Two Sigma Investments. \\

\clearpage
\newpage

\bibliographystyle{plainnat}
\bibliography{references}

\clearpage
\newpage
\appendix

\section{Implementation Details with Code Snippets}
\label{appendix:code_snippets}

\begin{figure}[h]
    \centering
    \includegraphics[width=125mm]{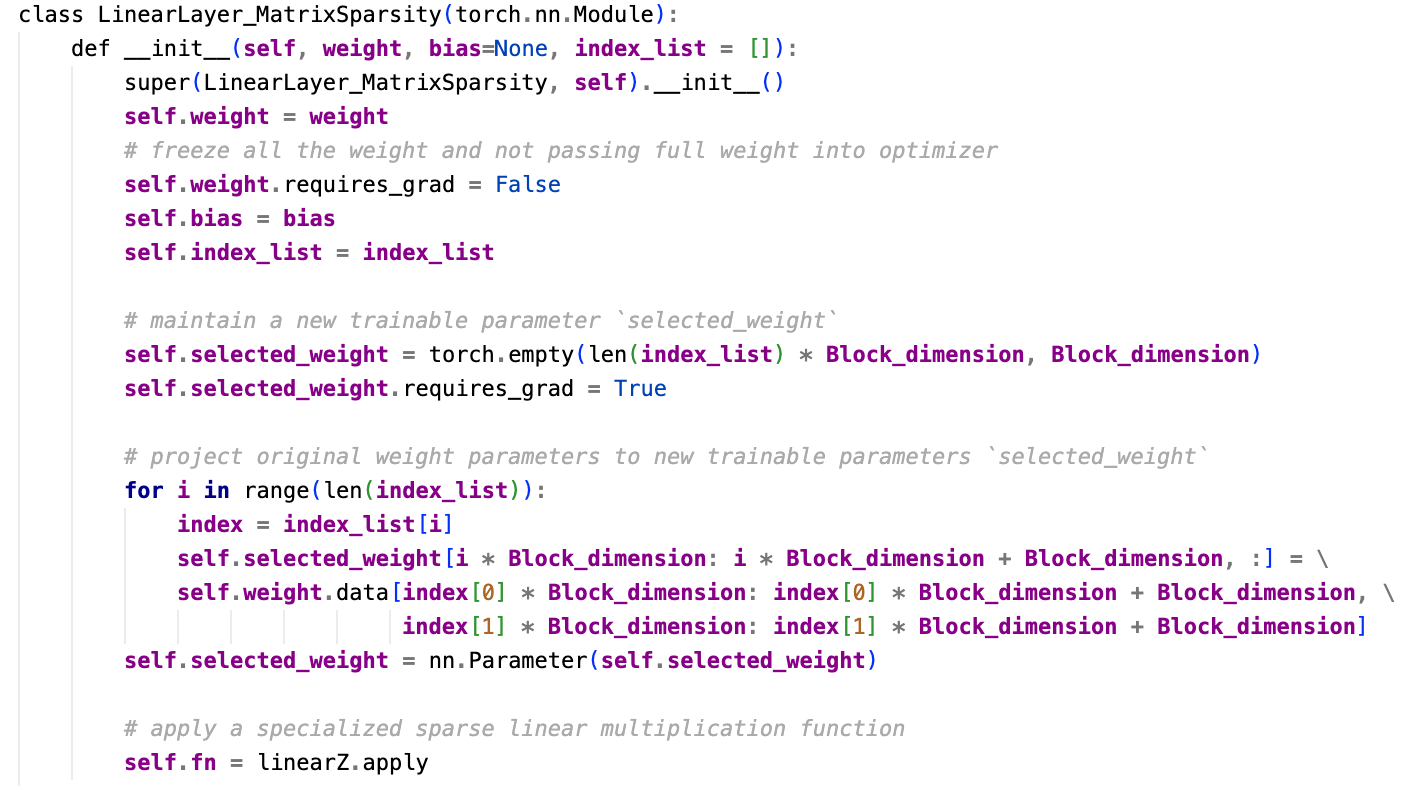}
    \centering
    \caption{Implementation of customized sparse linear layer.}
    \label{fig:code_sparse_linear_layer}
\end{figure}

\begin{figure}[h]
    \centering
    \includegraphics[width=125mm]{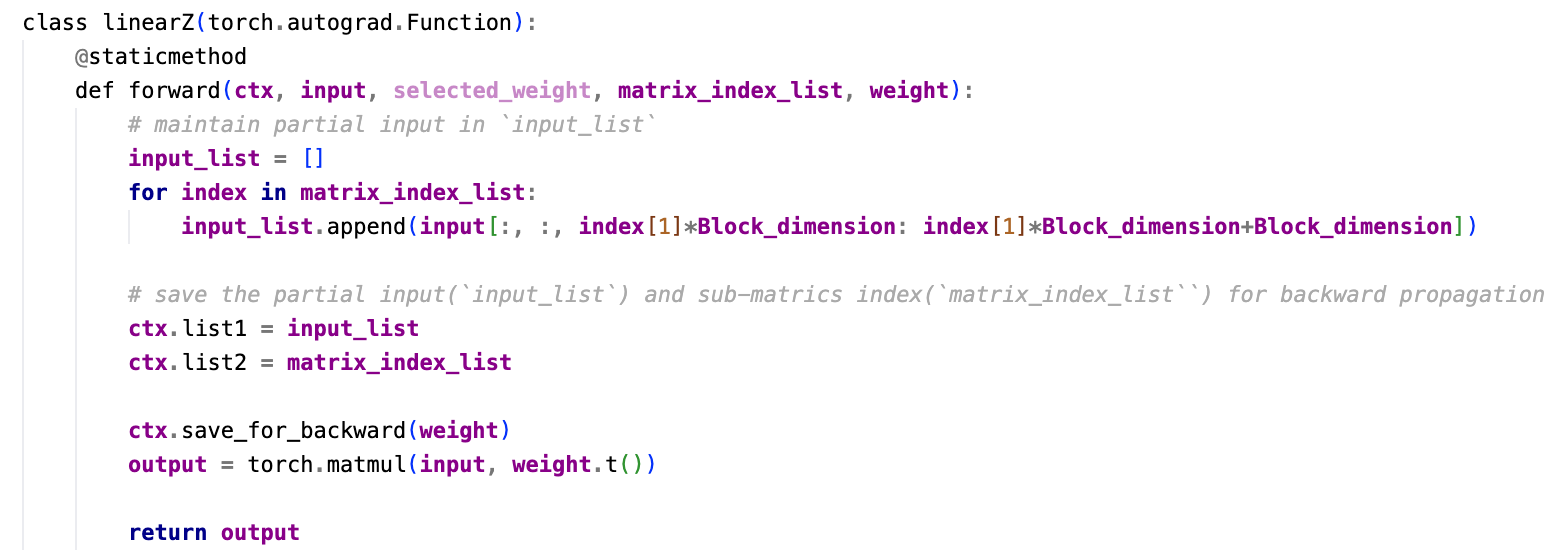}
    \centering
    \caption{Implementation of customized forward in specialized sparse linear multiplication function.}
    \label{fig:code_forward}
\end{figure}

\begin{figure}[h]
    \centering
    \includegraphics[width=125mm]{
    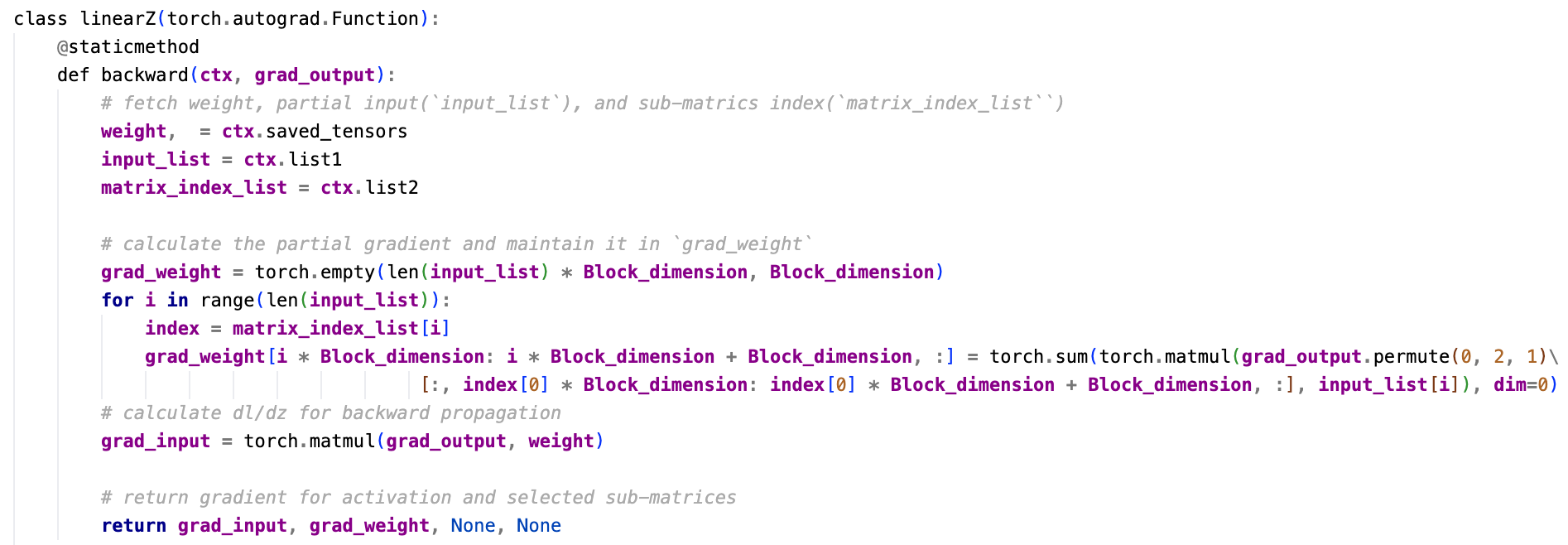}
    \centering
    \caption{Implementation of customized backward in specialized sparse linear multiplication function.}
    \label{fig:code_backward}
\end{figure}

\clearpage
\newpage
\section{Sub-Matrices Selection}
\label{appendix:sub_matrices_selection}
SMT ranks the average absolute gradient values within each sub-matrix and selects those with the highest averages. The rationale behind this selection process is to enable SMT to automatically identify sub-matrices containing memory that is most relevant to downstream tasks. During fine-tuning, the absolute gradient values can indicate the relevance of a block to these tasks hence it requires more tuning. By averaging the absolute gradient values within each sub-matrix, we can determine the importance of the sub-matrix to specific downstream tasks.

\section{Math10K Dataset}
\label{appendix:math10k}

\texttt{Math10K} dataset can evaluate the effectiveness of LLMs on the arithmetic reasoning task. \texttt{Math10K} incorporate six subsets including \texttt{GSM8k}, \texttt{SingleEq}, \texttt{SVAMP}, \texttt{MultiArith}, \texttt{AddSub}, and \texttt{AQuA}.(1) the \texttt{GSM8K} \cite{cobbe2021gsm8k} dataset consists of high quality linguistically diverse grade school math word problems created by human problem
writers, (2) the \texttt{SVAMP} \cite{patel-etal-2021-nlp} benchmark consists of one-unknown arithmetic word
problems for up-to-4 grade level students by making simple changes to a set of problems from another existing dataset, (3) the \texttt{MultiArith} \cite{roy2016solving} dataset of math word problems requiring multiple reasoning steps and operations, (4) the \texttt{AddSub} \cite{hosseini2014learning} dataset of addition and subtraction arithmetic word problems,
(5) the \texttt{AQuA} \cite{ling2017program} dataset of algebraic word problems with natural language rationales, and (6) the \texttt{SingleEq} \cite{koncel2015parsing} dataset of grade-school algebra word problems that map to single equations with varying length;

\end{document}